\pdfoutput=1

\documentclass[11pt]{article}

\usepackage[]{acl}

\usepackage{times}
\usepackage{latexsym}

\usepackage[T1]{fontenc}

\usepackage[utf8]{inputenc}

\usepackage{microtype}

\usepackage[inline]{enumitem}

\usepackage{tabularx}
\usepackage{booktabs}
\usepackage{tikz}
\usepackage{multirow}
\usepackage{amsmath}
\usepackage{amssymb}
\usepackage{longtable}

\definecolor{c0}{cmyk}{1,0.3968,0,0.2588} 
\definecolor{c1}{cmyk}{0,0.6175,0.8848,0.1490} 
\definecolor{c2}{cmyk}{0.1127,0.6690,0,0.4431} 
\definecolor{c3}{cmyk}{0.6765,0.2017,0,0.0667} 
\definecolor{c4}{cmyk}{0.3081,0,0.7209,0.3255} 
\definecolor{c5}{cmyk}{0,0.8765,0.7099,0.3647} 
\definecolor{darkgrey}{RGB}{180,180,180}
\definecolor{decentgrey}{RGB}{220,220,220}

\newcommand{\ours}{Unnatural Instructions}

\usepackage{makecell}
\usepackage{pifont}
\newcommand{\xmark}{\ding{55}}%

\title{Unnatural Instructions:\\Tuning Language Models with (Almost) No Human Labor}

\author{Or Honovich$^\tau$ \qquad Thomas Scialom$^\mu$ \qquad Omer Levy$^{\tau\mu}$ \qquad Timo Schick$^\mu$\\
$^\tau$ Tel Aviv University\\
$^\mu$ Meta AI}

\begin{document}
\maketitle
\begin{abstract}
Instruction tuning enables pretrained language models to perform new tasks from inference-time natural language descriptions.
These approaches rely on vast amounts of human supervision in the form of crowdsourced datasets or user interactions.
In this work, we introduce \emph{\ours{}}: a large dataset of creative and diverse instructions, collected with virtually no human labor.
We collect 64,000 examples by prompting a language model with three seed examples of instructions and eliciting a fourth.
This set is then expanded by prompting the model to rephrase each instruction, creating a total of approximately 240,000 examples of instructions, inputs, and outputs.
Experiments show that despite containing a fair amount of noise, training on \ours{} rivals the effectiveness of training on open-source manually-curated datasets, surpassing the performance of models such as T0++ and Tk-Instruct across various benchmarks.
These results demonstrate the potential of model-generated data as a cost-effective alternative to crowdsourcing for dataset expansion and diversification.\footnote{We make our data publicly available: \\\url{https://github.com/orhonovich/unnatural-instructions}}
\end{abstract}
\section{Introduction}

\definecolor{modelcolor}{RGB}{164,67,128}

\definecolor{promptcolor}{RGB}{0,0,0} 

\begin{figure}
    \centering
    \includegraphics[width=\linewidth]{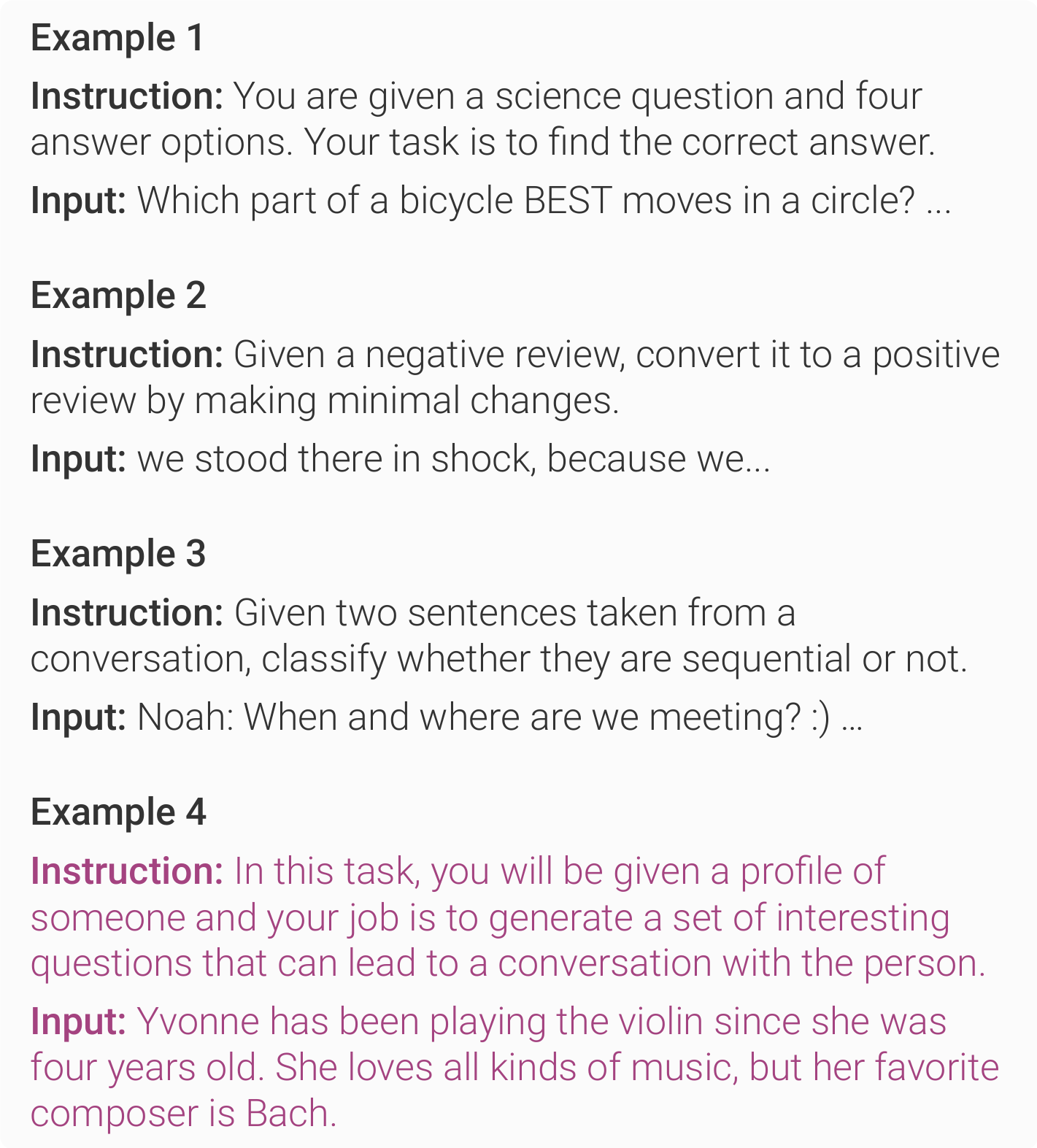}
    \caption{An illustration of our data generation prompt. \textbf{Black}: The prompt provided to the model. \textbf{\textcolor{modelcolor}{Pink}}: One of the model's generations for the given prompt. The full prompt is presented in Figure~\ref{fig:prompt}.}
    \label{fig:fig1_prompt}
    \vspace{-0.5cm}
\end{figure}
Instruction tuning enables pretrained language models to generalize to unseen tasks in a zero-shot setting \cite{sanh2021multitask, wei2021finetuned}.
One way to collect examples of instructions and their execution is to reformulate existing NLP datasets in an explicit instruction-input-output format via prompt engineering \cite{mishra-etal-2022-cross, supernaturalinstructions}.
However, the resulting data is limited to existing academic benchmarks, even though the instruction paradigm can describe any text-based task \cite{turkingtest}.
Alternatively, \citet{instructgpt} collect user-generated prompts and manually annotate their expected outputs, reflecting a different (and arguably more desirable) distribution of the instruction space, but requiring a live application with existing users and major investments in human annotation.
Can we create a large dataset of instructions that is diverse in tasks, content, and phrasing, \textit{without} human labor?

We introduce \textbf{\ours{}}, a dataset of natural language instructions and their corresponding inputs and outputs. 
Inspired by recent work on utilizing language models for data generation \citep{schick-schutze-2021-generating, lee2021neural, liu-etal-2022-wanli}, we collect data in a fully automatic manner by prompting a pretrained language model with three examples from the Super-Natural Instructions\footnote{Also known as Natural Instructions v2.} dataset \cite{mishra-etal-2022-cross, supernaturalinstructions} and asking the model to generate a fourth (Figure~\ref{fig:fig1_prompt}).
We repeat this process with 5 different seeds -- i.e. the entire process requires only 15 instruction examples -- to automatically produce 64,000 diverse triplets of instructions, inputs, and outputs.\footnote{In practice, we collected 68,478 examples, but only used subsets of 64,000 examples for training.}
We further diversify the dataset's format by generating additional natural language paraphrases of each instruction, while preserving the contents of any input arguments and outputs, expanding the dataset to approximately 240,000 examples.
Although the dataset contains noise, our analysis reveals that more than 50\% of generated examples are indeed correct, and that even incorrect examples typically contain valuable information for instruction tuning.
At the same time, we find that \ours{} contains highly creative tasks -- some of which are very different from ``classic'' NLP tasks -- and has a more diverse set of instructions than Super-Natural Instructions.

Experiments show that fine-tuning an 11B-parameter T5 model \cite{2020t5} on \ours{} can outperform both T0++ \cite{sanh2021multitask} and Tk-Instruct \cite{supernaturalinstructions} across several benchmarks, including Super-Natural Instructions \cite{supernaturalinstructions}, BIG-bench Hard \cite{bbh}, and LMentry \cite{lmentry}.
When controlling for all variables besides the data, we find that a model trained on \ours{} performs competitively with a baseline model trained on Super-Natural Instructions.
In particular, we observe an 18-point gain on BIG-bench Hard (original task formulation) and a 16-point gain on LMentry, suggesting that \ours{} is particularly useful for generalizing to instructions that deviate from the distribution of classic NLP tasks. 
These improvements become even more pronounced when the cost of generating examples is amortized; in this case, training on \ours{} substantially outperforms our baseline on  all benchmarks.
We observe a log-linear relationship between the number of generated examples and downstream task performance, suggesting that performance of models trained on \ours{} can further be improved simply by increasing its size.

Beyond the immediate implications on instruction tuning, this work demonstrates the viability of automatic dataset expansion using language models as an alternative to crowdsourcing.
\ours{} highlights the ability of language models to produce creative and diverse data, a trait that is difficult to obtain with crowd workers, who lack the intrinsic motivation to create novel examples and typically collapse into predictable heuristics to form annotation artifacts \cite{gururangan-etal-2018-annotation}.
At the same time, language models are faster and cheaper than human labor, opening up new possibilities for scaling up data annotation.

\section{Data Collection}
\label{sec:data_collection}

\definecolor{modelcolor}{RGB}{164,67,128}

\definecolor{promptcolor}{RGB}{0,138,212} 

\begin{figure*}
    \centering
    \includegraphics[width=\linewidth]{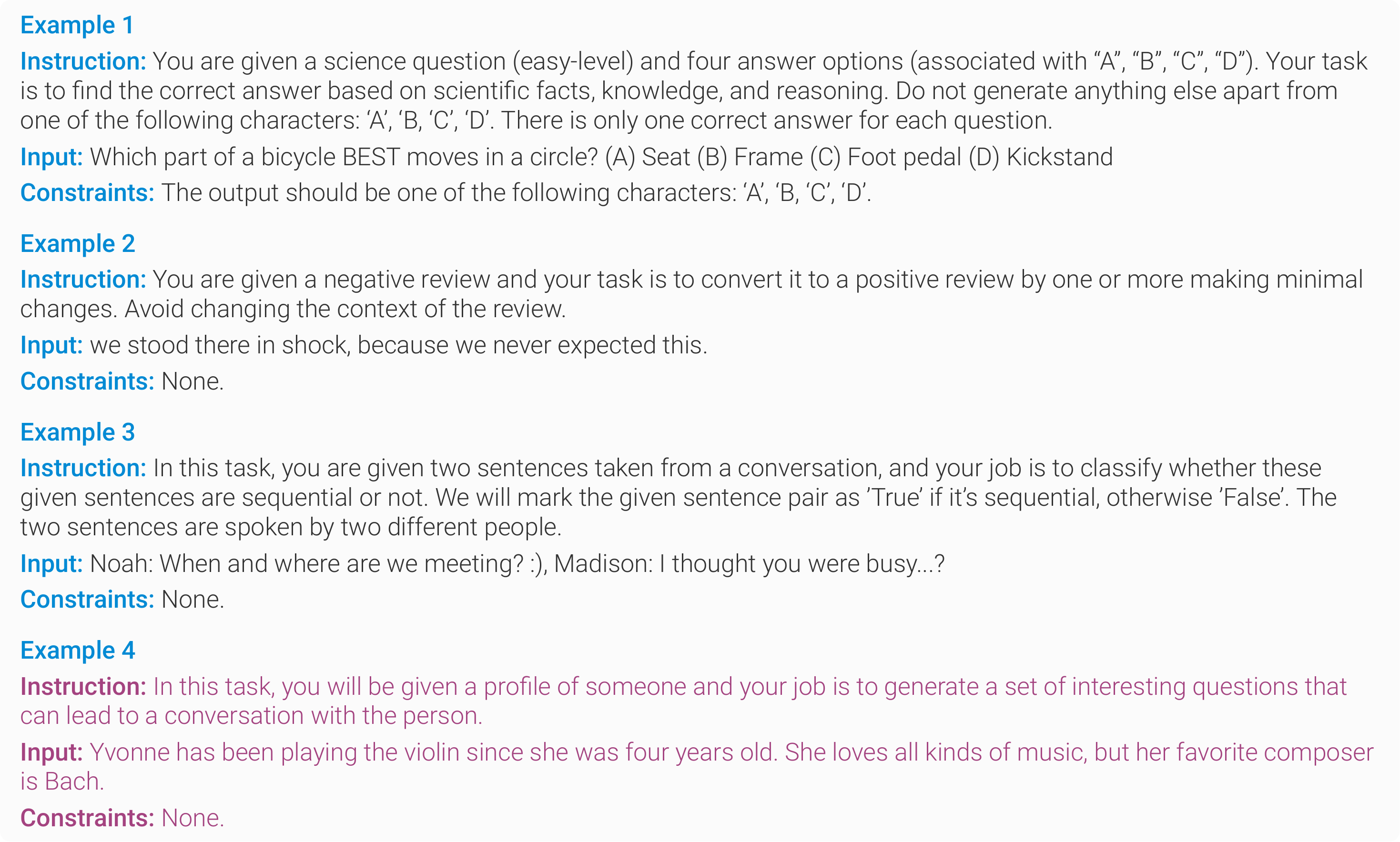}
    \caption{Our data generation prompt. \textcolor{promptcolor}{\textbf{Blue}}: The meta-prompt, which contains the number of the in-context example, as well as the constant fields of each example: instruction, input, and constraints. \textbf{Black}: The in-context examples. We show here one of our 5 in-context seeds. \textcolor{modelcolor}{\textbf{Pink}}: One of the model's generations for the given prompt. The generated example includes an instruction, input, and constraints.}
    \label{fig:prompt}
\end{figure*}

\begin{figure*}[t!]
\centering
 \includegraphics[width=1.0\textwidth]{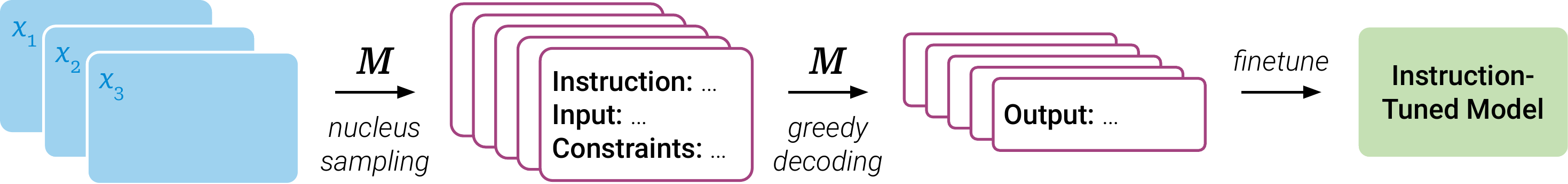}
 \caption{The core \ours{} generation pipeline. We use a seed of three in-context demonstrations $x_1, x_2, x_3$ to create a large dataset of NLP tasks with instructions, inputs and outputs. As a first step, we sample instructions, inputs, and constraints from a language model $M$. In the next step, we use $M$ to deterministically generate the corresponding outputs. Finally, the data can be used for instruction tuning.}
 \label{fig:high_level_process}
\end{figure*}

We introduce \ours{}, a dataset of 240,670 natural language instructions for a wide variety of natural language tasks.
Each example contains a natural language instruction as input and its expected execution as output.
Table~\ref{tab:analysis_tab} displays examples from the dataset.

\ours{} is collected in a completely automatic process, requiring a seed of only 15 manually-constructed examples, which can be produced in about one hour of human labor.
We first collect a core set of 68,478 examples (\S\ref{sec:core_generation}) by prompting a pretrained language model $M$\footnote{Throughout this section, we use OpenAI's text-davinci-002 as $M$. See \S\ref{sec:other_models} for experiments with other models.} with a seed of 3 manually-annotated examples to produce a new (fourth) example. This phase uses a structured instruction format and filtering heuristics to ensure data quality.
We then expand the core dataset by rephrasing the structured instructions in free-form natural language (\S\ref{sec:template_expansion}). This expansion is performed automatically by prompting a language model with manually-constructed examples, scaling up the dataset more than 3-fold.

\subsection{Core Dataset Generation}
\label{sec:core_generation}

The core dataset consists of examples in a structured format, making it easier for the generating model $M$ to predict and for us to filter automatically. We use stochastic decoding to generate example inputs (to promote creativity), followed by deterministic decoding to generate their outputs (for accuracy). Figure~\ref{fig:high_level_process} illustrates the process.

\paragraph{Format}
Each example in the core dataset contains four fields: 
\begin{enumerate}[label=(\arabic*)]
\item An \textbf{instruction} describing the task. The instruction can be a generic template (e.g. ``Write whether the following review is positive or negative'') that can be instantiated by a particular input argument (e.g. the review itself).
\item The \textbf{input} argument that instantiates the instruction, creating a specific example of the task.
\item Output space \textbf{constraints}, which detail the restrictions on the task's output space. Constraints are mainly relevant for classification tasks; for tasks with no specific output space constraints, this field is ``None.''
\item A textual \textbf{output} reflecting a correct execution of the instruction given the input arguments and output space constraints.
\end{enumerate}
The first three fields (instruction, input argument, constraints) are the model's input, and the output field acts as the reference for training and/or evaluation.
The constraints field is meant to guide $M$ during output generation and is discarded after generating the outputs (see next).
In \S\ref{sec:ablations} we provide data-driven evidence for selecting this particular format.

\paragraph{Input Generation}
The first step in the data generation pipeline is to generate examples of instruction-input-constraints.
We do so by prompting a model with three task demonstrations $x_1, x_2, x_3$, each presented in the structured instruction-input-constraint format (without outputs).
These demonstrations are wrapped by a simple meta-prompt that incentivizes the model to create a fourth example $x_4$, which we collect.
This process is illustrated in Figure~\ref{fig:prompt}.

We use 5 different seeds of 3 demonstrations each to generate the entire core dataset. In other words, the whole process requires only 15 manually-constructed examples.
All demonstrations are taken from the Super-Natural Instructions \citep{supernaturalinstructions} train set.
To obtain various examples using the same prompt, decoding is done by nucleus sampling (top $p$) with $p=0.99$ \cite{Holtzman2020The}.

\paragraph{Filtering}
We apply three automatic filters to the generated examples to remove: (1) model generations that do not include the three input fields (instruction, input argument, and constraints), (2) instructions and inputs that are identical to those demonstrated in the prompt, (3) duplicate examples, i.e. two different examples that have the same instruction and input argument.

\paragraph{Output Generation}
Given a generated example $x$, we generate the corresponding output $y$ by conditioning a pretrained language model with the instruction, input argument, and constraints (if not none), followed by an ``Output:'' prompt.
Here we apply greedy decoding to prioritize correctness over creativity. We ignore examples for which the generated output is an empty string.

\subsection{Template Expansion}
\label{sec:template_expansion}
Examples in the \ours{} core dataset have a strict instruction-input-output format.
To increase the format diversity and obtain tasks phrased in free-form natural language \cite{schick-schutze-2021-shot, sanh2021multitask}, we collect alternative formulations that preserve the content of the original instructions.
Specifically, we prompt a language model to reformulate the tasks in the core dataset and collect two alternative formulations for each generated task.\footnote{The seed reformulations in each prompt are inspired and partially taken from PromptSource \citep{bach-etal-2022-promptsource}.}
The alternative formulations are often shorter and less formal than the original instructions.
The rephrasing prompt contains two examples of instructions and their alternative formulation.
We do not include inputs, constraints, and outputs in the rephrasing prompt; instead, we utilize the already-generated inputs and outputs to complement the rephrased instruction.
Unlike the examples in the core dataset, in some alternative formulations, the input is embedded into the task description rather than following it.
We achieve that by adding an ``\{INPUT\}'' placeholder, which marks the position for input insertion (Figure~\ref{fig:rephrasing_prompt}).

In some cases, the model generates two identical additional formulations, while in others, it copies the original instruction. Some alternative formulations may also have an invalid format - e.g., not containing the ``\{INPUT\}'' placeholder. When such failures occur we continue to sample alternative formulations, stopping after five unsuccessful attempts. For this reason, some instructions have only one alternative formulation, while others have none. Overall, more than 97.5\% of the instructions have two valid and distinct alternative formulations.

In fact, some instructions end up with more than two paraphrases because we generate two paraphrases per \emph{example} (i.e. instruction-input-output pair) and the core dataset contains examples that share the exact same instruction but not the same input argument. Therefore, by cross-referencing each instruction's alternative phrasings with all of its input arguments, we can extend the data even further and arrive at a total of 240,670 examples without additional cost.

\definecolor{modelcolor}{RGB}{164,67,128}

\definecolor{promptcolor}{RGB}{0,138,212} 

\begin{figure}
    \centering
    \includegraphics[width=\linewidth]{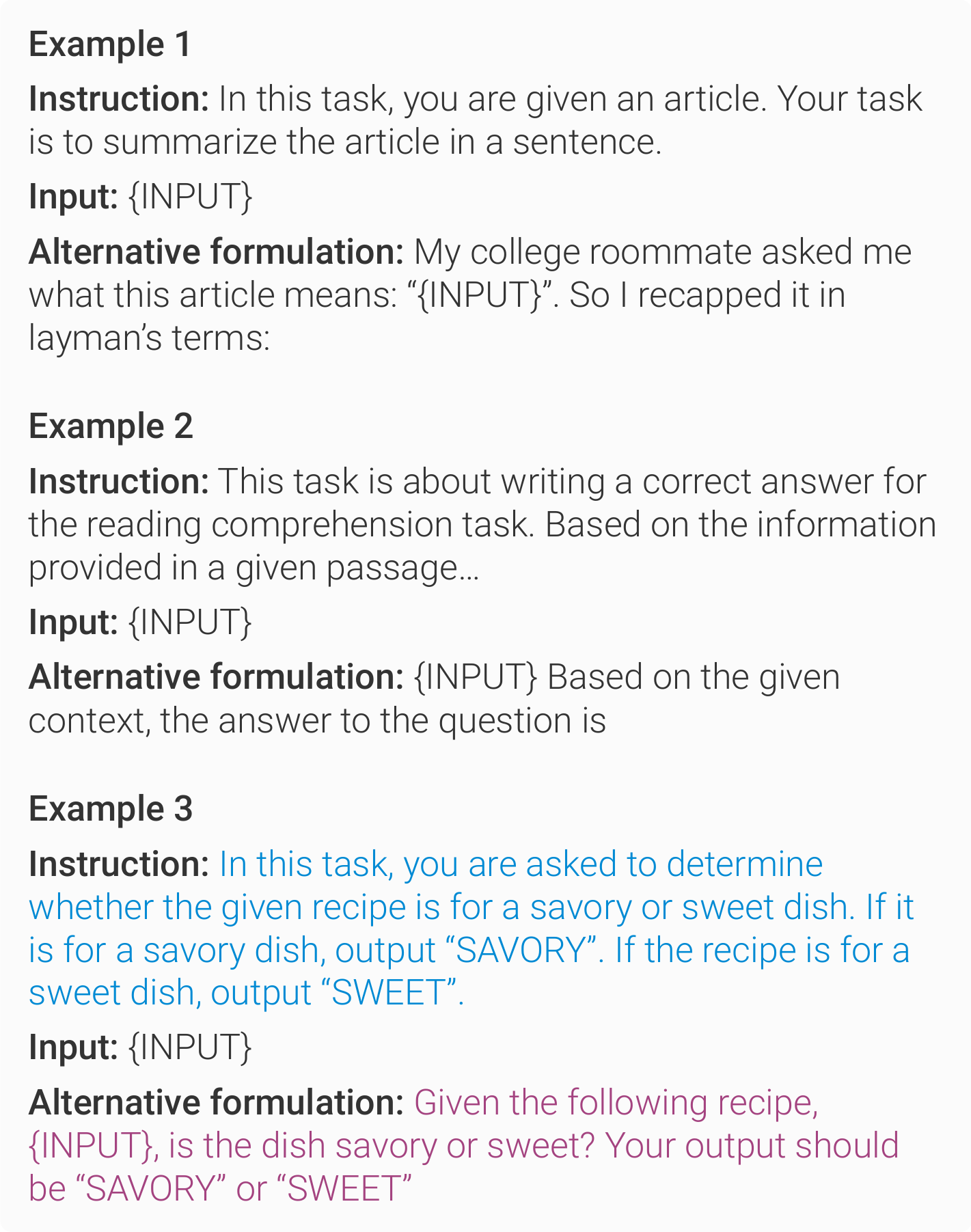}
    \caption{Our template expansion prompt. \textbf{Black}: Few-shot demonstrations of instructions and a possible alternative formulation. \textcolor{promptcolor}{\textbf{Blue}}: A model-generated instruction for which the model should suggest an alternative formulation. \textcolor{modelcolor}{\textbf{Pink}}: An example of a model-generated task reformulation.}
    \label{fig:rephrasing_prompt}
    \vspace{-0.5cm}
\end{figure}
\section{Data Analysis}
\label{sec:analysis}

\begin{table*}[t]
    \small
    \centering
    \begin{tabular}{@{}p{0.75\textwidth}p{0.2\textwidth}@{}}
    \toprule
    \textbf{Instruction} & \textbf{Category}\\
    \midrule
    You need to answer the question 'Is this a good experiment design?', given an experiment scenario. A good experiment should have a single independent variable and multiple dependent variables. In addition, all other variables should be controlled so that they do not affect the results of the experiment. & Experiment Verification \\
    \midrule
    You are given a recipe for baking muffins that contains some errors. Your task is to correct the errors in the instructions by replacing each underlined word with the correct one from the options provided. & Recipe Correction \\
    \midrule
    You will be given a piece of text that contains characters, places, and objects. For each character in the text, you need to determine whether they are static or dynamic. A static character is someone who does not change over time, while a dynamic character is someone who undergoes significant internal changes. & Character Categorization \\
    \midrule 
    In this task, you are asked to generate a limerick given two rhyming words. A limerick is a five-line poem with the following rhyme scheme: AABBA. The first, second and fifth lines must be of three beats, while the third and fourth lines must be of two beats each. Additionally, all poems should have the same meter (e.g., iambic pentameter) & Poem Generation \\
    \midrule
    I'm not sure what this idiom means: ``\{INPUT\}''. Could you give me an example? & Idiom Explanation \\
    \midrule
    \{INPUT\} By analyzing the writing styles of the two passages, do you think they were written by the same author? & Author Classification \\
    \midrule
    I need to invent a new word by combining parts of the following words: \{INPUT\}. In what order should I put the parts together? & Word Invention \\
    \midrule
    What is the punchline to the following joke? \{INPUT\} & Humor Understanding \\
    \bottomrule
    \end{tabular}
    \caption{Examples of eight interesting generated instructions and their corresponding category. The first four examples are taken from the core dataset, while the last four were generated during the template expansion phase.}
    \label{tab:creativity_tab}
\end{table*}

We first demonstrate the \emph{creativity} of instructions in \ours{} and then manually analyze 200 examples, randomly sampled from our core dataset, focusing on \textit{correctness} and \textit{diversity}. We also compare the distribution of \ours{} to that of Super-Natural Instructions, and find that the inputs of \ours{} tend to be more diverse.

\paragraph{Creativity}
A major challenge when creating a general-purpose instructions dataset is task creativity. Crowd workers may struggle to do so, and typically collapse into predictable heuristics to form annotation artifacts \cite{gururangan-etal-2018-annotation}. While the high performance of models trained on \ours{} across several benchmarks (see \S\ref{sec:experiments}) suggests that it is indeed diverse and creative, we additionally present in Table~\ref{tab:creativity_tab} some cherry-picked examples of the generated instructions, providing a glimpse at their creativity.

\begin{table*}[t]
    \small
    \centering
    \begin{tabularx}{\linewidth}{@{}Xc@{}}
    \toprule
    \textbf{Example} & \textbf{Correct?}\\
    \midrule
     \makecell[X]{\textbf{Instruction}: In this task, you are asked to come up with a design for an office lobby. The client has provided specific instructions on what they want in the lobby, and it is your job to create a floor plan that meets all of their requirements.\\ \vspace{-5pt} \textbf{Input:} Client wants a welcoming environment with space for people to wait comfortably without feeling cramped. They also would like some sort of area where people can do work if they need to kill some time before their next meeting. \\ \vspace{-5pt} \textbf{Output:} [...] Based on these requirements, a possible floor plan [...] The lobby would have a reception area [...] with access to outlets, and a coffee station.} & $\checkmark$ \\
    \midrule
     \makecell[X]{\textbf{Instruction}: You will be given several pieces of information about an event, and you have to determine whether or not it is a cause-and-effect relationship. If the given statements are related by cause and effect, then output `True'. Otherwise, output `False'.\\ \vspace{-5pt} \textbf{Input:} Statements: [`The tornado damaged the city', `Many people were left homeless']. \\ \vspace{-5pt} \textbf{Output:} True} & $\checkmark$ \\
    \midrule
     \makecell[X]{\textbf{Instruction}: You are to determine if the statement is true or false.\\ \vspace{-5pt} \textbf{Input:} Text: Santa Claus lives at the North Pole. \\ \vspace{-5pt} \textbf{Output:} False} & \xmark \\
    \midrule
     \makecell[X]{\textbf{Instruction}: You are given a scientific experiment scenario and your job is to determine which variable(s) the scientist should control in order to test the hypothesis.\\ \vspace{-5pt} \textbf{Input:} The hypothesis is ``If a plant receives more sunlight, then it will grow taller.'' Which variable(s) should the scientist control? \\ \vspace{-5pt} \textbf{Output:} A} & \xmark \\   
    \bottomrule
    \end{tabularx}
    \caption{Examples of generated instructions, inputs and outputs in our core dataset. For the first two examples, the entire pair of instruction, input and output is valid. The third example has an incorrect output; in the fourth example, the experiment is not described in the input.}
    \label{tab:analysis_tab}
\end{table*}

\paragraph{Correctness}
When evaluating correctness, we test whether (1) the generated instructions are logical and executable, (2) the input arguments correspond to the task described in the instruction, and (3) the outputs are correct, given the instruction and input. 
Although our data filtering process is minimal, 113 of the 200 analyzed examples (56.5\%) are correct.
Of the 87 incorrect examples, 9 (4.5\%) had incomprehensible instructions, 35 (17.5\%) had an input that did not match the task description, and 43 (21.5\%) had incorrect outputs.
Table \ref{tab:analysis_tab} shows some correct and incorrect examples from our analysis.

While the amount of noise in the data may raise concerns regarding its usability, we note that many of the examples that were marked as incorrect can still be considered informative. For example, one erroneous example had the instruction \textit{``In this task, you will be provided with a list of countries and their corresponding capital cities. You are also given a list of clues to help you solve the puzzle. For each clue, determine which country it is referring to and write down that country's name in the space next to the clue...''}
The input argument was \textit{``Clue 1: This capital city is on two different continents.''}
This example was marked as incorrect since the input did not conform with the format described by the instruction -- a list of countries and their capitals was not provided, only a clue. However, the output was \textit{Istanbul, Turkey}, which indeed lies in both Europe and Asia and therefore corresponds with the clue provided as input.
In \S\ref{sec:experiments} we provide quantitative evidence that, despite being noisy, Unnatural Instructions provides a highly informative training signal.

\paragraph{Diversity}
We manually cluster the instructions into task types and measure the number of unique task types. Out of the 200 examples tested, we identify 117 distinct tasks. While many tasks are classical NLP tasks, such as sentiment analysis, question answering, and summarization, others are not quite canonical, and some are very specific, such as detecting a recipe given a list of ingredients. Table~\ref{tab:most_common_instructions} shows the most commonly generated tasks from the set of 200 analyzed examples. Other tasks appeared 3 times or less, with 85 tasks appearing only once.

\begin{table}[t]
    \small
    \centering
    \begin{tabular}{@{}lr@{}}
    \toprule
    \textbf{Task} & \textbf{\#Examples}\\
    \midrule
    Question Answering & 11 \\
    Sentiment Analysis & 10 \\
    Arithmetic & 8 \\
    Geometry & 8 \\
    Event Ordering & 7 \\
    Fact Verification & 5 \\
    Fill-in-the-Blank & 5 \\
    General Math Puzzles & 4 \\
    Identifying Overlapping Strings & 4 \\
    Array Manipulations and Puzzles & 4 \\
    \bottomrule
    \end{tabular}
    \caption{Top 10 tasks by \#examples, out of the 200 manually-analyzed examples from the core dataset of \ours{}.}
    \label{tab:most_common_instructions}
    \vspace{-0.5cm}
\end{table}

\begin{figure}
    \centering
    \includegraphics[width=\linewidth]{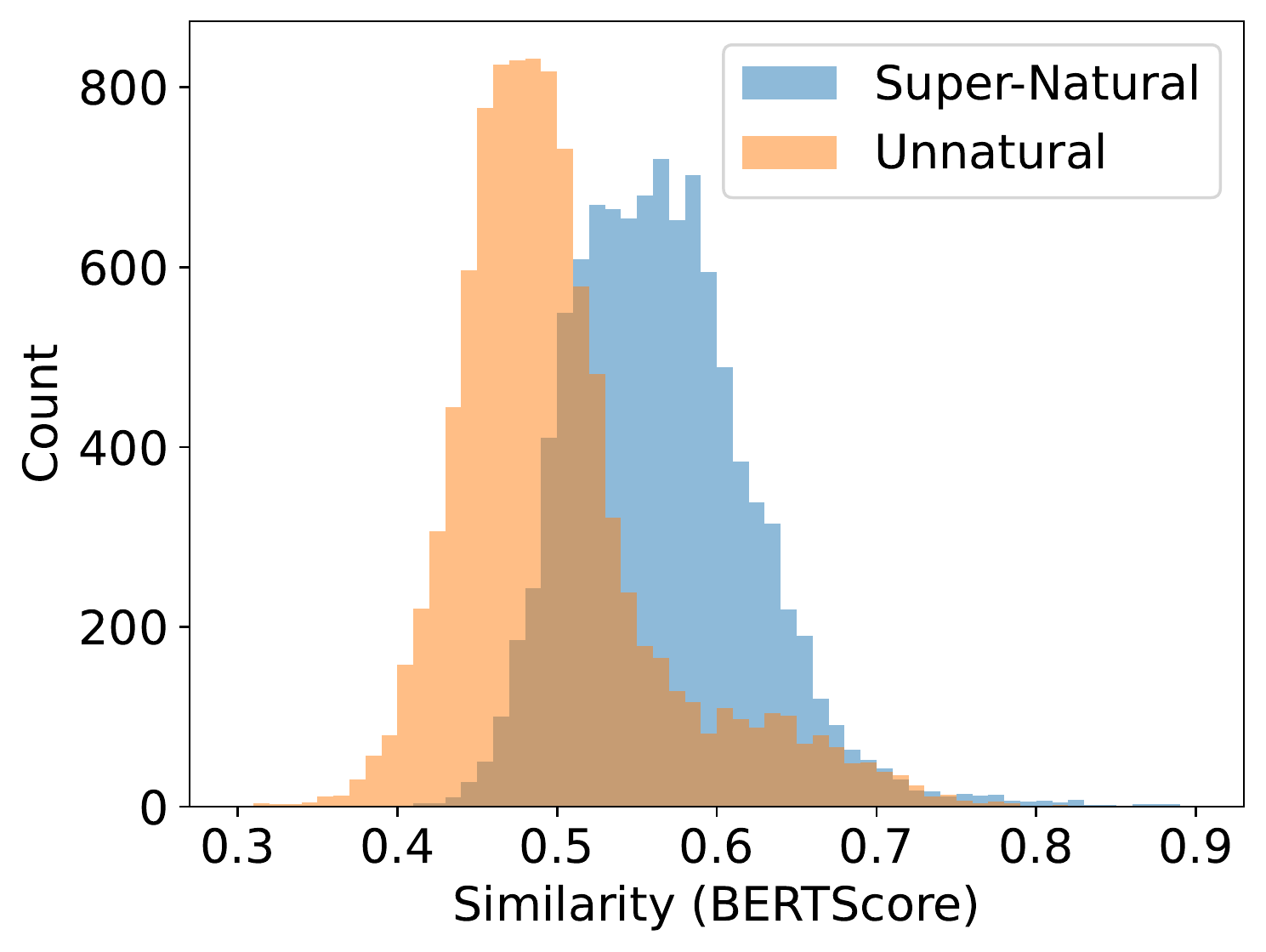}
    \caption{Similarity scores distribution for Super-Natural Instructions and for \ours{}, obtained by sampling 10,000 pairs of examples from each dataset and computing their similarity.}
    \label{fig:diversity}
    \vspace{-0.25cm}
\end{figure}

We also analyze how similar each pair of examples is, as a general proxy for diversity. Specifically, we sample 10,000 pairs of examples from \ours{}, and compute the similarity of their inputs using BERTScore \cite{bertscore}.\footnote{Using the deberta-xlarge-mnli model \cite{he2021deberta}.}
We repeat this process for Super-Natural Instructions, producing two empirical distributions.
Figure~\ref{fig:diversity} shows that the inputs of \ours{} tend to be less similar to each other than the inputs of Super-Natural Instructions.
This result comes as a surprise considering the fact that the entire \ours{} dataset was constructed by conditioning only on 15 original examples.

\section{Experimental Setup}
\label{sec:setup}

We describe how we use \ours{} to fine-tune models and elaborate on our evaluation protocol.

\subsection{Fine-Tuning on \ours{}}

We fine-tune T5-LM, the language-model-adapted variant of T5-11B \citep{2020t5, lester-etal-2021-power}.
We follow standard practice for fine-tuning, using a batch size of 16 examples over 3 epochs. For training on our core dataset, we use the same template as \citet{supernaturalinstructions} for formatting instructions and inputs.
Our full set of training hyperparameters is available in Appendix~\ref{sec:hyperparameters}.
We create a small validation set of 1,000 examples for model selection following the methodology proposed by \citet{supernaturalinstructions}: we randomly select 10 examples from 100 random tasks of the Super-Natural Instructions training set.

\subsection{Baselines}

We measure the relative utility of \ours{} by comparing it to a variety of models, all based on T5-11B, which were fine-tuned with different types and quantities of manually-annotated instruction data.

\paragraph{T0++}
\citep{sanh2021multitask} is an instruction-tuned variant of T5-LM, trained on tasks in the PromptSource \citep{bach-etal-2022-promptsource} prompt formats.

\paragraph{Tk-Instruct}
\citet{supernaturalinstructions} fine-tune T5 v1.1 on Super-Natural Instructions, using a subsample of 757 tasks with 100 examples each.
Tk-Instruct is trained with a batch size of 1,024 examples for 1,000 steps.
Since our evaluation focuses on zero-shot instruction understanding, we use the definition-only version of Tk-Instruct.

\paragraph{FLAN-T5}
\citet{flampalm} fine-tune T5 on a collection of tasks phrased as instructions in multiple prompting setups (zero-shot, few-shot, Chain-of-Thought \citep{wei2022chain}), achieving impressive zero-shot generalization capabilities.

\paragraph{T5-LM on Natural Instructions}
Our main point of comparison is the utility of the original manually-curated instructions in Super-Natural Instructions.
We therefore train a model which is identical to ours in all aspects but data.
Specifically, we fine-tune the LM-adapted variant of T5-11B on a subsample of 64,000 examples from Super-Natural Instructions training set, excluding examples from any task that participates in the validation set.
This model differs from Tk-Instruct along three aspects: the dataset subsample, the base model (T5-LM), and some training hyperparameters (batch size 16 for 3 epochs).

\subsection{Evaluation}

We evaluate models on four different benchmarks, measuring a range of capabilities.
All evaluations are carried out in a zero-shot setting, without few-shot demonstrations, unless explicitly provided in the instructions.

\paragraph{Natural Instructions}
We evaluate models on the test set of Super-Natural Instructions \citep{mishra-etal-2022-cross, supernaturalinstructions}. As in the original papers, outputs are generated using greedy decoding, and performance is measured using Rouge-L.

\paragraph{T0: Zero-Shot}
We evaluate models on the held-out set of T0 \citep{sanh2021multitask}, using rank classification for decoding and accuracy as a metric. For fair comparison, we remove tasks supersets of which are present in the Tk-Instruct training set. The final set contains six tasks: ANLI R1-R3, CB, COPA and RTE. We refer to this evaluation set as T0: Zero-Shot.
Unlike Super-Natural Instructions, T0: Zero-Shot tasks do not have a strict format and are phrased in a rather free-form manner, including inputs that can be embedded into the task description. We therefore expect models trained on our core dataset (without instruction paraphrases) to perform poorly under these conditions, while adding the task reformulation data should boost performance on T0: Zero-Shot.

\paragraph{BIG-bench: Hard}
The ``hard'' subset of BIG-bench \citep{bbh} contains 23 challenging tasks from BIG-Bench \cite{srivastava2022beyond}. We investigate two different formats for all tasks: their original format in BIG-bench, and the format of \citet{bbh}, who reformulate each task as question answering with manually added instructions; for the latter, we remove all few-shot demonstrations. For both formats, we use greedy decoding and exact match with the reference for evaluation.

\begin{table*}[t]
\small
\centering
\begin{tabular}{@{}lrcccc@{}}
\toprule
\multirow{2}{*}{\textbf{Model}}  & \multirow{2}{*}{\textbf{\#Examples}} & \textbf{Super-Natural}  &  \textbf{T0:} &   \textbf{BIG-bench:} & \multirow{2}{*}{\textbf{LMentry}} \\
  & & \textbf{Instructions}  &  \textbf{Zero-Shot} & \textbf{Hard (Orig/QA)} & \\
\midrule
\textbf{\textit{Prior Work}} & & & & & \\
T5-LM & 0 & 24.3  & 40.2  & \phantom{0}0.0 / \phantom{0}0.7 & 20.6 \\
T0++ & 12,492,800 & 40.3  &  \textsc{nho}  & 20.2 / 13.9 & 38.3 \\
Tk-Instruct & 75,417 & 45.6  &  41.4  & \phantom{0}5.8 / 11.8 & 35.7 \\
FLAN-T5 & 14,336,000 & \textsc{nho} & \textsc{nho} & \underline{39.3} / \underline{40.0} & \underline{52.2} \\
\midrule
\textbf{\textit{Direct Comparison Baseline}} & & & & & \\
T5-LM on Super-Natural Instructions & 64,000 &  \underline{\textbf{54.0}}  &  44.0  & 10.2 / \textbf{29.7} & 34.6 \\
\textbf{\textit{Our Approach}} & & & & & \\
T5-LM on \ours{} & 64,000 &  51.9  &  45.7 & 16.0 / 29.5 & 42.0 \\
\qquad $+$ Instruction Paraphrases  & 240,670 &  49.3 & \underline{\textbf{49.0}} & \textbf{28.1} / 29.4 & \textbf{50.7} \\
\bottomrule
\end{tabular}
\caption{Performance of several models on the four benchmarks considered. Best results in our direct comparison setup are bold, best results overall are underlined. \textsc{nho} indicates that a benchmark's data is not held out because it was used for training. T5-LM on \ours{} performs better than several strong baselines and is competitive to our direct comparison baseline, outperforming it in three setups despite being finetuned on automatically generated data only. Template expansion substantially increases performance in most cases but gives worse results on Super-Natural Instructions.}
\label{tab:main_results}
\end{table*}

\paragraph{LMentry}
LMentry \citep{lmentry} is a benchmark that tests basic language abilities, designed to complement common approaches for evaluating large language models. Outputs are generated by applying greedy decoding and evaluated using high-accuracy regular expressions. The benchmark's metric is the LMentry score, which combines accuracy with multiple aspects of robustness.

\section{Results}
\label{sec:experiments}

Our main results are shown in Table~\ref{tab:main_results}, which reports the performance of each model on each benchmark considered. 
Remarkably, T5-LM finetuned on \ours{} clearly outperforms several strong instruction-tuned baselines such as T0++ and Tk-Instruct; the only exception to this is BIG-bench: Hard (Orig), where T0++ performs better. Retraining a model on Super-Natural Instructions using our exact setup reveals that a much stronger performance than that of Tk-Instruct can be achieved using this dataset. However, even in this direct comparison setup, \ours{} leads to stronger or equal performance for every dataset except Super-Natural Instructions itself. While T5-LM finetuned on \ours{} is outperformed by FLAN-T5, the amount of training data for this model is larger by several orders of magnitude.
These results demonstrate that fully automated data generation with pretrained LMs is indeed a viable and cost-effective alternative to human-curated data.

\subsection{Performance with Template Expansion}
\label{sec:full-data-res}
We evaluate the contribution of template expansion (§\ref{sec:template_expansion}) to the performance of models trained on \ours. To this end, we finetune a single model on our full dataset with paraphrases; results are shown in the bottom row of Table~\ref{tab:main_results}.

Adding instruction paraphrases boosts performance on T0: Zero-Shot (+3.3), Big-bench: Hard in its original format (+12.1) and LMentry (+8.7).
We surmise that this improvement is largely because examples in our core dataset were generated based on demonstrations from Super-Natural Instructions only and therefore have their exact format and style. Accordingly, models trained on our core dataset rely too much on this specific format and cannot generalize well to different formats found in other benchmarks. Obtaining more format diversity through template expansion successfully addresses this issue. On the other hand, over-reliance on the format of Super-Natural Instructions is probably preferable when testing on this dataset itself, which explains the performance drop when adding paraphrases compared to the boost in performance on other benchmarks.

While some of the performance gains observed may also be attributed to the fact that adding paraphrases simply increases the data, in~\S\ref{sec:scaling_data} we show that template expansion is helpful even when controlling for dataset size.

\begin{figure*}[t]
    \centering
    \includegraphics[height=0.75\columnwidth]{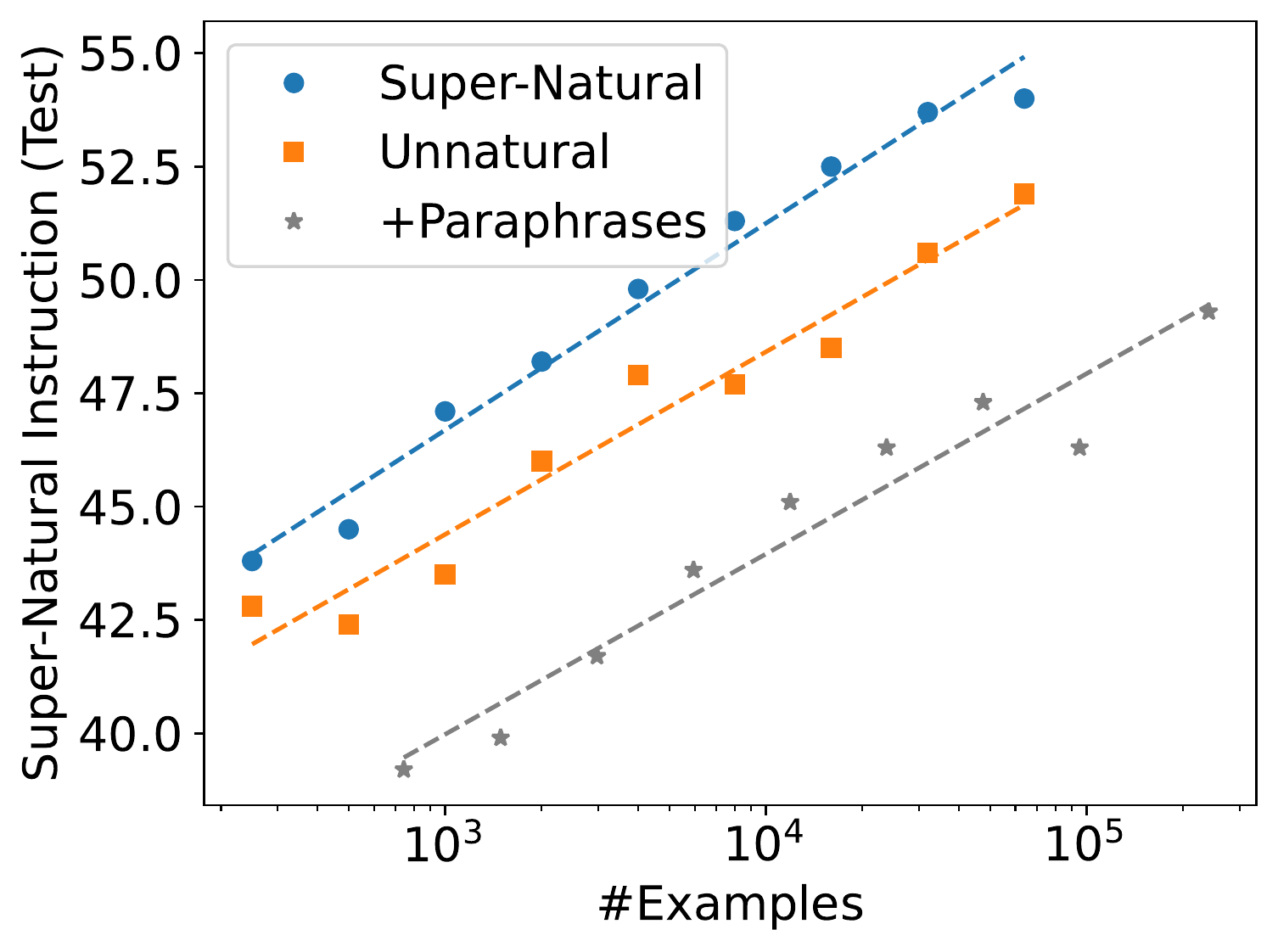} ~~~~ \includegraphics[height=0.75\columnwidth]{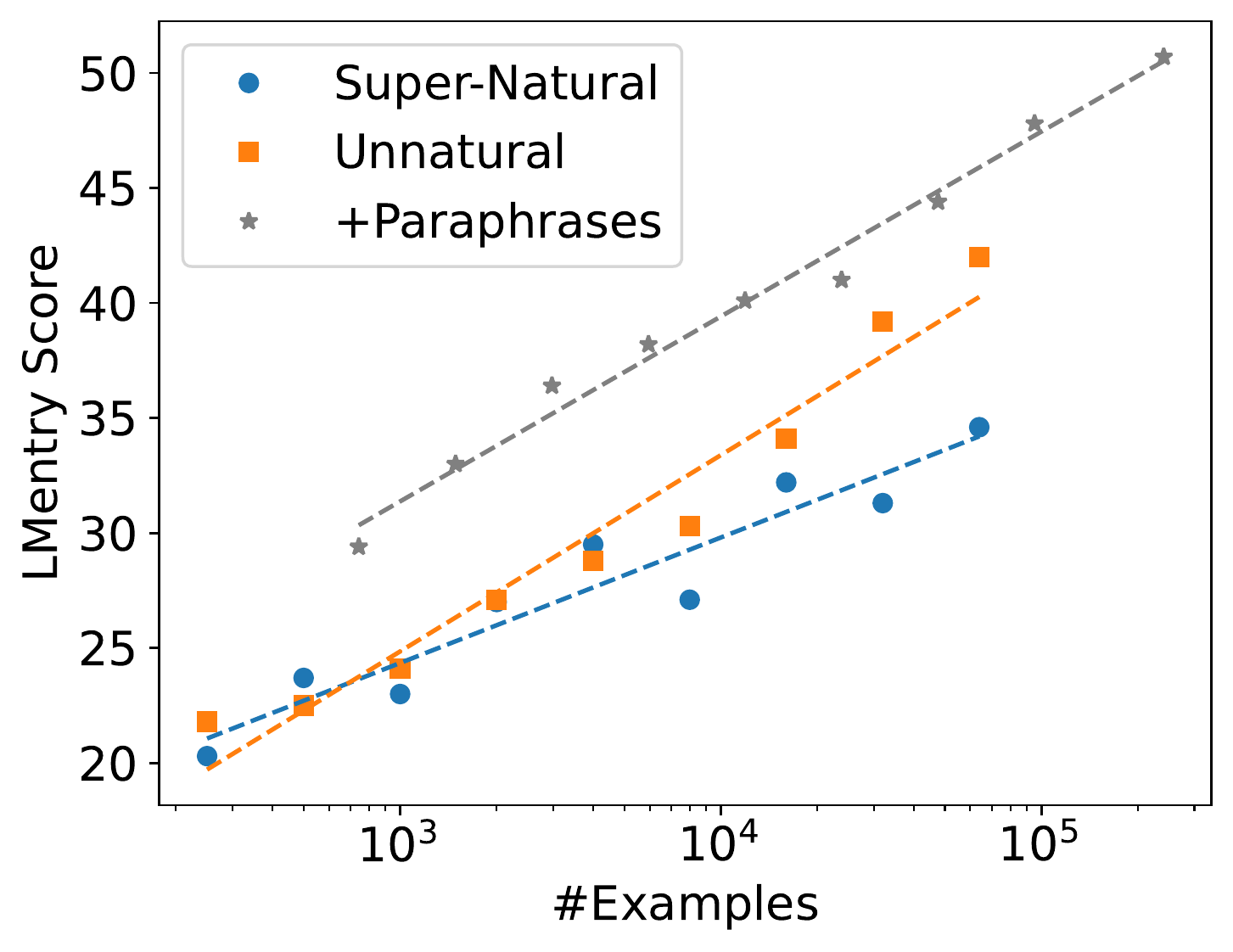}
    \vskip10pt
    \includegraphics[height=0.75\columnwidth]{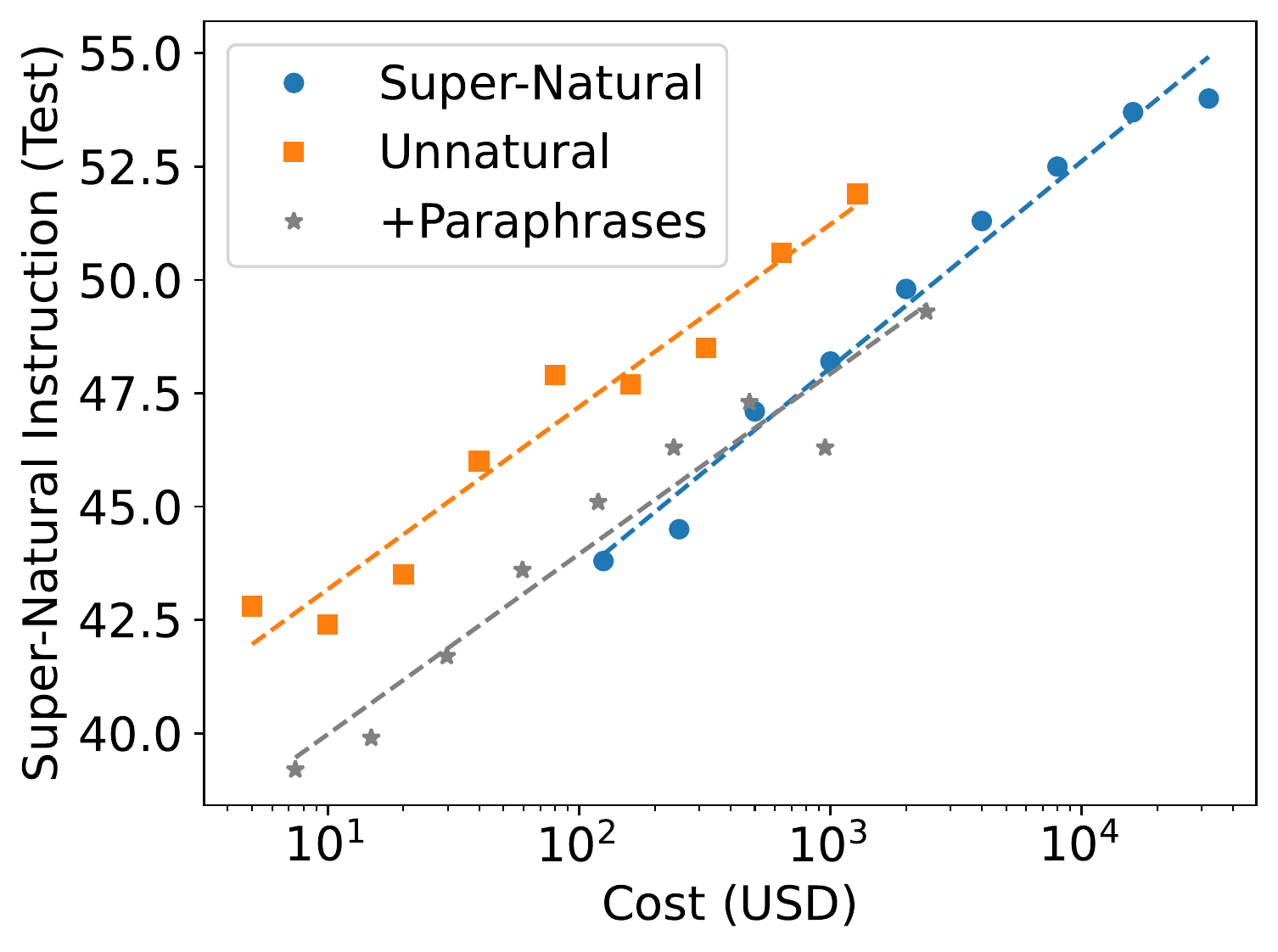} ~~~~ \includegraphics[height=0.75\columnwidth]{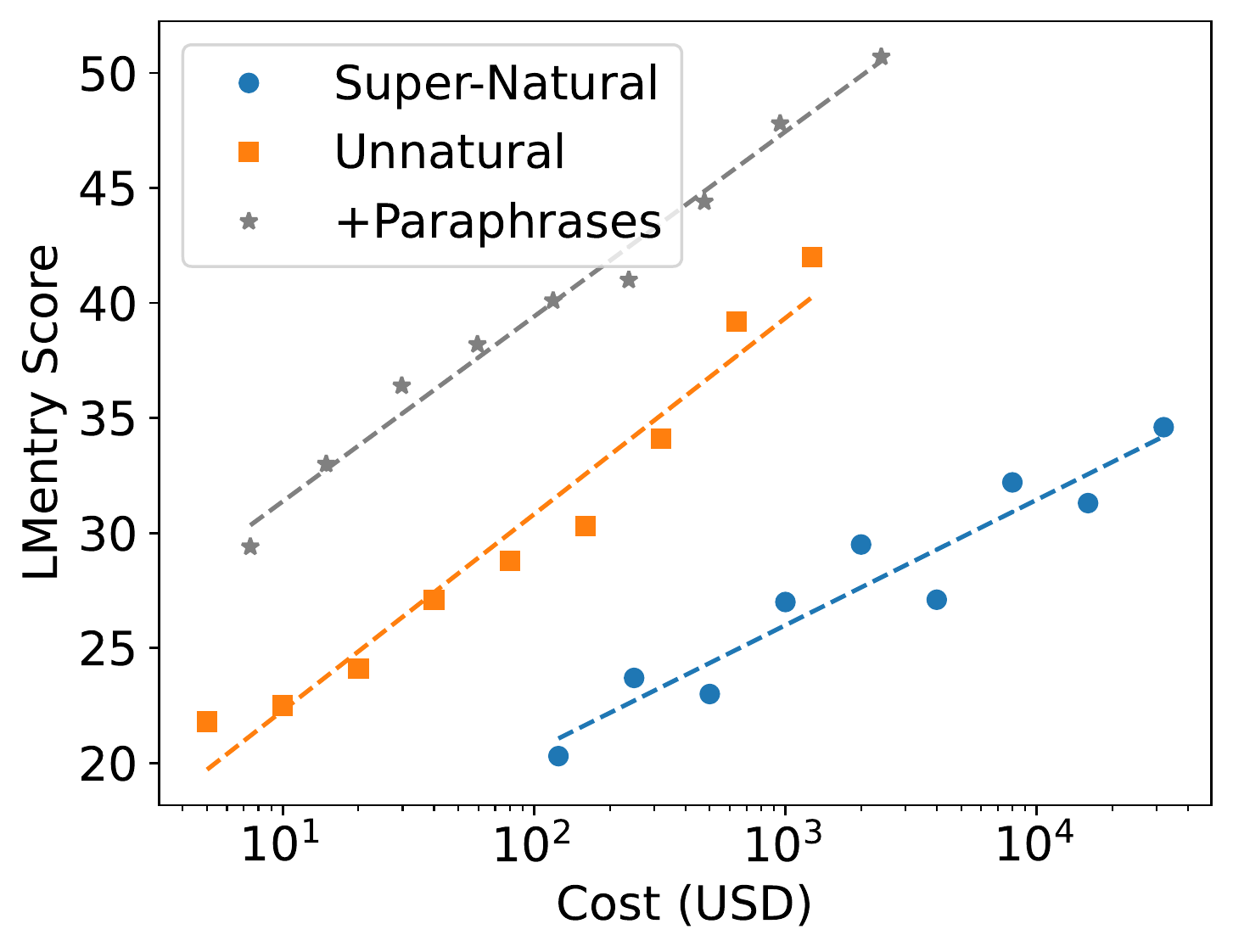}
    \caption{Scaling experiments comparing \ours{} with Super-Natural Instructions. \textbf{Top row:}~Model performance when controlling for \emph{dataset size}, tested on Super-Natural Instructions (left) and LMentry (right). \textbf{Bottom row:} Model performance when controlling for the \emph{cost of obtaining data}, tested on Super-Natural Instructions (left) and LMentry (right).}
    \label{fig:scale}
\end{figure*}

\subsection{Performance Scaling by Dataset Size}
\label{sec:scaling_data}

As all of our data is generated from the same model using the same set of prompts, scaling up the amount of generated examples might lead to numerous repetitions and, as a consequence, diminishing returns in terms of downstream task performance. To investigate whether this is an issue, we analyze how the amount of training examples affects the performance of our finetuned models. To this end, we train models on subsets of both Super-Natural Instructions and \ours{}, ranging from 250 to 64,000 examples. As shown in Figure~\ref{fig:scale} (top row), our core and full data as well as Super-Natural Instructions all exhibit log-linear scaling laws, suggesting that even for subsets of \ours{} containing thousands of examples, simply generating more examples still adds a valuable signal to our training data.

Results for LMentry (Figure~\ref{fig:scale}, top right) show that our template expansion process is still beneficial when controlling for dataset size. The added value of the paraphrases is therefore likely to be in terms of format diversity rather than solely as a method for increasing the amount of data.

\subsection{Performance Scaling by Cost}

In practical scenarios with fixed annotation budgets, the actual \emph{cost} associated with a certain level of performance is even more relevant than the number of required examples. We therefore measure model performance as a function of the cost for obtaining the training data. Based on OpenAI's pricing as of December 2022, the cost for generating an example is estimated at \$0.02 for our core dataset, and \$0.01 for the expanded dataset. \citet{kiela-etal-2021-dynabench} estimate human annotation cost at \$0.50--\$1.00 per example, excluding indirect costs such as task design and UX development; for comparison with our automatic data collection method, we assume the lower-bound human annotation cost of \$0.50.

As shown in Figure~\ref{fig:scale} (bottom row), \ours{} is clearly more cost-efficient than manually curated data. This is true even for the Super-Natural Instructions test set, where a model trained on \ours{} is weaker than a model trained on Super-Natural Instructions for a fixed number of examples, but better when controlling for cost, showing that our automatic data generation approach outperforms crowdsourcing for a fixed annotation budget.
\section{Data Collection Ablations}
\label{sec:ablations}

We explore the effect of the different components of our data collection pipeline by conducting structural prompt ablations. Throughout this section, we train models for 1,500 steps using 2,000 examples and evaluate the Super-Natural Instructions validation set performance, averaged across three different random seeds.

\subsection{Generative Model}
\label{sec:other_models}

As a data generation model, we used text-davinci-002, an instruction-tuned variant of GPT-3 \citep{gpt3}.
However, our approach is not limited to this specific model.
We experiment with generating examples using the original (untuned) GPT-3 model by using it as the model $M$ in both the input generation and output generation phases (see \S\ref{sec:data_collection}).

Table~\ref{tab:model-ablations} shows how replacing an instruction-tuned model with a vanilla model affects the quality of the data using performance on the Super-Natural Instructions validation set as a proxy.
We observe that while the quality of generated \emph{inputs} does drop by 4.5 points, it is well within the range of other prompt ablations (see the remainder of this section). In other words, informative and diverse \textit{instructions} can be generated by untuned language models.
However, generating \textit{outputs} does seem to require some level of instruction tuning. A manual analysis reveals that outputs generated by GPT-3 mainly suffer from the model's inability to stop (i.e. predict EOS), often starting with the correct answer, but then degenerating into repetitions or tangents.
While this may be remedied through various post-processing heuristics, we leave exploration of such methods to future work.

\begin{table}
\small
\centering
\begin{tabular}{@{}llr@{}} 
\toprule
\multicolumn{2}{@{}l}{\textbf{Model Used to Generate}}    & \textbf{Super-Natural}           \\
\textbf{Input} & \textbf{Output}     & \textbf{Instructions}           \\
\midrule
text-davinci-002 & text-davinci-002                & \textbf{48.7} $\pm$ 0.3 \\
GPT-3 & text-davinci-002                & 44.2 $\pm$ 0.7 \\
GPT-3 & GPT-3                & \phantom{0}4.1  $\pm$ 0.1 \\
\bottomrule
\end{tabular}
\caption{Performance of 11B T5-LM models trained on 2,000 examples, generated with different models, on the Super-Natural Instructions validation set.
}
\label{tab:model-ablations}
\end{table}

\subsection{Meta-Prompts}
\label{sec:metaprompts_ablation}

\begin{figure}
    \centering
    \includegraphics[width=\linewidth]{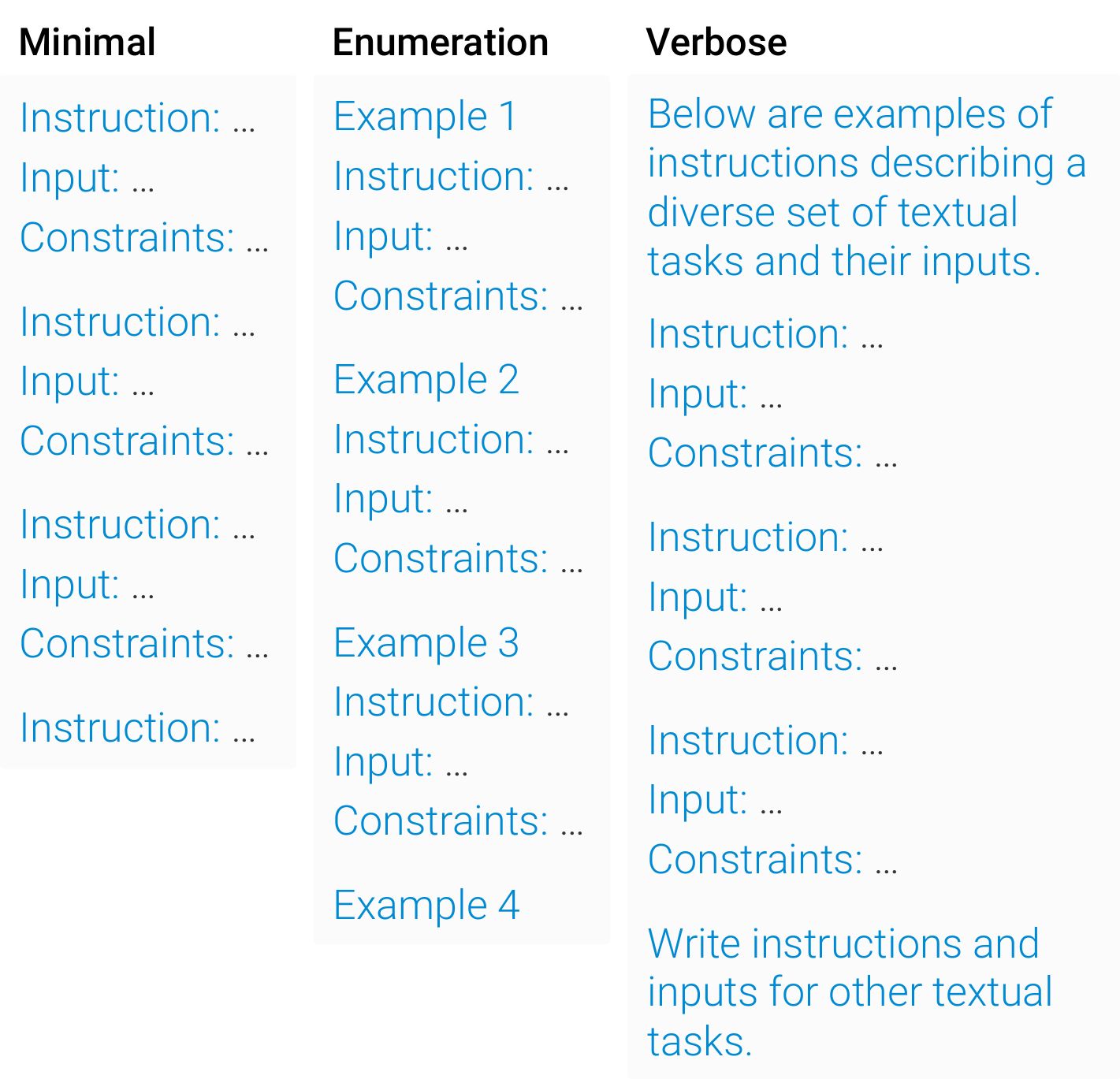}
    \caption{The meta-prompts used in our ablations.}
    \label{fig:short_metaprompts}
\end{figure}

Language models are known to be sensitive to the \textit{meta-prompt} -- i.e., the text wrapping the in-context demonstrations, which can include task description or additional guidance regarding the desired output.
We therefore experiment with three different meta-prompt styles: \emph{minimal}, \emph{enumeration}, and \emph{verbose} (Figure~\ref{fig:short_metaprompts}).

Table~\ref{tab:metaprompt-ablations} presents the results obtained from fine-tuning on datasets generated with different meta-prompts.
We observe that the simple enumeration approach elicits more informative examples than either the minimalistic or verbose approaches. Perhaps surprisingly, the verbose meta-prompt performs worse than the minimalistic one, possibly because the last line (the command) interrupts the pattern, and does not align well with patterns in the pretraining corpus.\footnote{While our core dataset was created using the enumeration meta-prompt, the remaining ablation experiments in this section were run using the verbose meta-prompt.}

\begin{table}[t]
    \small
    \centering
    \begin{tabular}{@{}lr@{}}
    \toprule
    \textbf{Meta-Prompt}  & \textbf{Super-Natural Instructions} \\
    \midrule
    Minimal & 47.5 $\pm$ 0.6\\
    Enumeration & \textbf{48.7 $\pm$ 0.3} \\
    Verbose & 46.9 $\pm$ 0.3 \\
    \bottomrule
    \end{tabular}
    \caption{Performance of 11B T5-LM models trained on 2,000 examples, generated with each meta-prompt, on the Super-Natural Instructions validation set.}
    \label{tab:metaprompt-ablations}
\end{table}

\subsection{In-Context Examples}

Models such as GPT-3 are known to be sensitive to slight variations in prompt content, resulting in performance differences when provided with different demonstrations sampled from the same dataset \citep{liu-etal-2022-makes} and when permuting the in-context demonstrations \citep{kumar-talukdar-2021-reordering,lu-etal-2022-fantastically}.
To account for the effect of the provided demonstrations on the quality of the generated data, we experiment with each of our five demonstration sets separately.\footnote{See Appendix~\ref{sec:prompts} for all demonstration sets.} 
Table~\ref{table:ice-ablations} shows that the data generation pipeline is largely robust to variations in the in-context demonstrations, with one outlier (seed 4).
Inspecting the differences between these groups, we find that seed 4 led to less constrained instructions: 1,376 out of 2,000 examples do not have constraints, whereas that number is between 28 and 880 for all other sets.
Indeed, in seed 4, only one out of three prompt demonstrations had constraints, while in other sets, at least two demonstrations had constraints.

\begin{table}
\small
\centering
\begin{tabular}{@{}lr@{}} 
\toprule
  \textbf{Seed Demonstrations} & \textbf{Super-Natural Instructions} \\
\midrule
1   & 46.9 $\pm$ 0.3 \\
2   & 46.1 $\pm$ 0.3 \\
3   & 46.8 $\pm$ 0.4 \\
4   & 41.9 $\pm$ 1.0 \\
5   & 46.0 $\pm$ 0.2 \\
\midrule
Mix & 46.1 $\pm$ 0.3 \\ 
\bottomrule
\end{tabular}
\caption{Performance of 11B T5-LM models trained on 2,000 examples, generated with various sets of three in-context demonstrations (seeds), on the Super-Natural Instructions validation set. \textit{Mix} samples 400 examples from each of the five single-seed datasets.}
\label{table:ice-ablations}
\end{table}

\subsection{Constraints}

As mentioned in \S\ref{sec:data_collection}, each instruction-input demonstration is accompanied by an additional \textit{constraints} field, which details the task's output space restrictions (e.g., ``entailment'', ``contradiction'' or ``neutral'' for NLI).
We note that, in all demonstrations, the instruction itself lists the output space constraints.
We hypothesize that adding the constraints field may emphasize these restrictions, ultimately steering the output generation model to produce outputs in the correct format.
We verify our hypothesis by conducting two ablation experiments.
First, we keep the constraints field when generating the instructions and inputs, but only use instructions and input arguments for the output generation step (i.e., without concatenating generated constraints).
Second, we completely remove the constraints field from the data generation pipeline, leaving the instruction field as the only source of information for output space constraints.
Table~\ref{table:constraint-ablations} shows that the constraints field has a positive effect both on the quality of the generated outputs \textit{and} inputs.
Removing constraints from the output generation step reduces performance by 3 points, and removing the field from the instructions-inputs generation phase decreases performance by an additional 2.2 points.

\begin{table}
\small
\centering
\begin{tabular}{@{}ccc@{}} 
\toprule
\multicolumn{2}{@{}c}{\textbf{Use ``Constraints:'' for}} & \textbf{Super-Natural}\\
\textbf{Input Gen} & \textbf{Output Gen}                & \textbf{Instructions}           \\
\midrule
$\checkmark$ & $\checkmark$                         & \textbf{46.9 $\pm$ 0.3}  \\
$\checkmark$ &      & 43.9 $\pm$ 0.7           \\
 &        & 41.7 $\pm$ 0.2           \\
\bottomrule
\end{tabular}
\caption{Performance of 11B T5-LM models trained on 2,000 examples, generated with and without the \textit{constraints} field, on the Super-Natural Instructions validation set.}
\label{table:constraint-ablations}
\end{table}

\subsection{Two-Step Process}

An alternative to our two-step pipeline is to generate instruction-input-output triplets in one pass. To test this approach, we provide the model with the same prompt used for the instruction-input-constraints generation, only with an additional \textit{output} field, added after the constraints field. As Table~\ref{table:constraint-ablations} shows, one-step generation obtains a score that is lower by 1.7 than the default two-step process.
We suspect that this gap is a result of using stochastic decoding in the unified input-output generation phase, which is critical for obtaining diverse inputs. In contrast, when generating outputs in a separate phase, we can use deterministic decoding algorithms to maximize accuracy.

\begin{table}
\small
\centering
\begin{tabular}{@{}lr@{}} 
\toprule
\textbf{Data Generation Process}                 & \textbf{Super-Natural Instructions}           \\
\midrule
Separate I/O Steps     & \textbf{46.9 $\pm$ 0.3}  \\
Unified I/O Step       & 45.2 $\pm$ 0.6           \\
\bottomrule
\end{tabular}
\caption{Performance of 11B T5-LM models trained on 2,000 examples, generated either using separate input and output steps or a single unified step, on the Super-Natural Instructions validation set.}
\label{table:steps-ablations}
\end{table}

\section{Related Work}

\paragraph{Instruction Tuning}
\citet{turkingtest} propose the Instruction Paradigm, where models learn new tasks from natural language instructions alone. \citet{mishra-etal-2022-cross, supernaturalinstructions} construct the first large-scale instruction benchmarks by collecting crowdsourcing instructions used to create NLP datasets and converting them into a uniform format.
\citet{sanh2021multitask, wei2021finetuned} further extend the usability of instructions by suggesting \textit{instruction tuning}, where a language model is trained on many natural language instructions in the hope that it will generalize to new, unseen instruction tasks. \citet{flampalm} advance instruction tuning by scaling the number of tasks, scaling the model size, and adding chain-of-thought \citep{wei2022chain}, while \citet{instructgpt} propose a reinforcement learning approach for instruction tuning from comparative human judgements.

\paragraph{Automatic Data Generation}
Obtaining large-scale supervised data can be expensive and time-consuming.
To mitigate this, several studies have explored automatic data generation.
A common approach is to automatically augment existing datasets  \citep[\textit{inter alia}]{AnabyTavor2020DoNH, andreas-2020-good, yang-etal-2020-generative, Kaushik2020Learning, lee2021neural}. \citet{kiela-etal-2021-dynabench} suggest a human-and-model-in-the-loop dataset creation, where a model is trained on initial data, then annotators are asked to seek examples that are misclassified by the model, in an iterative process. In the same manner, \citet{nie-etal-2020-adversarial} apply a process to create training data for the task of NLI \citep{dagan-pascal-2003, bowman-etal-2015-large}, obtaining state-of-the-art performance on a variety of NLI benchmarks. 
\citet{liu-etal-2022-wanli} combine human annotators and GPT-3 to create challenging examples for NLI.

While all the above techniques require an existing labeled dataset, other work suggested creating datasets entirely automatically, without the need for labeled data. \citet{schick-schutze-2021-generating} propose to leverage pretrained language models to generate entire datasets of labeled text pairs from scratch. \citet{agrawal2022qameleon} use pretrained language models to automatically construct multilingual QA data using only five examples per language.
To the best of our knowledge, \ours{} is the first work to go beyond a particular task and automatically generate a large-scale general-purpose dataset, which emphasizes task diversity.

\section{Conclusion}
We introduce \ours{}, an automatically generated dataset of natural language instructions and their corresponding inputs and outputs. To the best of our knowledge, this is the first general-purpose NLP dataset that was automatically generated. Our experiments show that models trained on \ours{} can outperform models trained on manually annotated datasets across several benchmarks. \ours{} is not only very cost-effective, we also provide evidence of enhanced diversity in the instructions produced and a high level of creativity in the tasks devised, a trait difficult to obtain with crowd workers.
Ablations show that even weaker models without instruction tuning can generate useful instructions, though they may struggle with producing the corresponding outputs. However, coming up with interesting tasks and writing diverse instructions for them is arguably the main challenge of the data collection process, whereas given instructions and inputs, outputs are often far easier to annotate through crowdsourcing. Our findings incentivize utilizing models for general-purpose data generation, which we view as an intriguing direction for future research.

\bibliography{anthology, custom, galileo}

\newpage
\appendix
\section{Fine-Tuning Hyperparameters}
\label{sec:hyperparameters}

We use the same set of hyperparameters for fine-tuning experiments with T5-LM \citep{2020t5, lester-etal-2021-power}. All models are trained for up to $\max(3\text{ epochs}, 3000\text{ steps})$ and the final model is chosen based on Rouge-L on our validation set, where we evaluate every 100 steps. We use a batch size of 16, a maximum learning rate of $1\cdot10^{-5}$ with warm-up for the first 10\% of training and a weight decay of $0.01$. We truncate inputs at 1,024 tokens and outputs at 128 tokens. All models are trained using DeepSpeed's ZeRO-3 \citep{10.1145/3394486.3406703}. Training on up to 64,000 examples is performed on 32 NVIDIA Tesla V100 16GB Volta GPUs using FP32; for bigger training datasets, we used 8 NVIDIA A100 40GB GPUs with BF16.
For computing Rouge-L and exact match scores, we use the implementation of \citet{supernaturalinstructions}.

\section{Evaluation Details}

For evaluating model performance on Super-Natural Instructions, T0: Zero-Shot and LMEntry, we use their official evaluation scripts. For evaluation on BIG-bench: Hard, we lowercase outputs, remove punctuation characters and trim extra whitespace before computing exact match scores. The only exception to this is the task \texttt{dyck\_languages}, where the target output consists entirely of punctuation characters.

\section{Data Generation Prompts}
\label{sec:prompts}
Table~\ref{table:incotext-prompts-tab} presents the in-context demonstrations we used, taken from \citet{supernaturalinstructions}.

\onecolumn

\newpage
\definecolor{promptcolor}{RGB}{0,138,212} 

\begingroup
\small
\begin{longtable}{@{}p{\textwidth}@{}}\\
\toprule
\textbf{In-Context Demonstrations} \\
\midrule
\textbf{Seed 1} \\*
\\*
\textcolor{promptcolor}{Example 1}\\*
\textcolor{promptcolor}{Instruction: }In this task, you're given passages that contain mentions of names of people, places, or things. Some of these mentions refer to the same person, place, or thing. Your job is to write questions that evaluate one's understanding of such references. Good questions are expected to link pronouns (she, her, him, his, their, etc.) or other mentions to people, places, or things to which they may refer. Do not ask questions that can be answered correctly without understanding the paragraph or having multiple answers. Avoid questions that do not link phrases referring to the same entity. For each of your questions, the answer should be one or more phrases in the paragraph, and it should be unambiguous.\\*
\textcolor{promptcolor}{Input: }Passage:  Nearing London, Oliver encounters Jack Dawkins, a pickpocket more commonly known by the nickname the "Artful Dodger", and his sidekick, a boy of a humorous nature named Charley Bates, but Oliver's innocent and trusting nature fails to see any dishonesty in their actions. The Dodger provides Oliver with a free meal and tells him of a gentleman in London who will "give him lodgings for nothing, and never ask for change". Grateful for the unexpected assistance, Oliver follows the Dodger to the "old gentleman's" residence. In this way Oliver unwittingly falls in with an infamous Jewish criminal known as Fagin, the gentleman of whom the Artful Dodger spoke. Ensnared, Oliver lives with Fagin and his gang of juvenile pickpockets in their lair at Saffron Hill for some time, unaware of their criminal occupations. He believes they make wallets and handkerchiefs.\\*
\textcolor{promptcolor}{Constraints: }None.\\*
\\*
\textcolor{promptcolor}{Example 2}\\*
\textcolor{promptcolor}{Instruction: }You will be given a piece of text either about an everyday event, or a general statement. If the event seems a plausible event to you, or the general statement makes sense matches your commonsense, output 'True', otherwise output 'False'.\\*
\textcolor{promptcolor}{Input: }Text: The glass fell of a three-story building, so it broke into pieces.\\*
\textcolor{promptcolor}{Constraints: }The output should be one of the two: `True' or `False'.\\*
\\*
\textcolor{promptcolor}{Example 3}\\*
\textcolor{promptcolor}{Instruction: }You need to answer the question 'Are the given steps in order?', given a set of steps describing a process. Your answer must be either Yes or No. If the answer is No, that means the steps are out of order and do not make sense in the order they are in. If the answer is Yes, that means the steps are in order and make sense in the order that they are in. A set of steps are not in order if the steps reference information that is introduced in a later step.\\*
\textcolor{promptcolor}{Input: }Steps: [`The seeds are dispersed by wind, animals, etc', `The seeds reach the ground', `Grow into new trees', `The process repeats itself over and over', `A tree produces seeds',`These new trees produce seeds']\\*
\textcolor{promptcolor}{Constraints: }The output should be one of the two: `Yes' or `No'.\\*
\\*  
\textcolor{promptcolor}{Example 4} \\*
\midrule
\textbf{Seed 2} \\*
\\*
\textcolor{promptcolor}{Example 1}\\*
\textcolor{promptcolor}{Instruction: }In this task, you are given two phrases: Head and Tail, separated with <sep>. The Head and the Tail events are short phrases possibly involving participants. The names of specific people have been replaced by generic words (e.g., PersonX, PersonY, PersonZ). PersonX is always the subject of the event. You have to determine whether the Head is used for the Tail or not. The usage describes everyday affordances or uses of objects and includes both typical and atypical uses. For example, a popcorn bucket can typically be used to hold popcorn, but it could also serve as a hat in atypical situations. Classify your answers into ``Yes'' and ``No''. The phrase may also contain ``-'', a placeholder that can be an object, a person, and/or an action.\\*
\textcolor{promptcolor}{Input: } Head: floor mats<sep>Tail: wipe off one's boots\\*
\textcolor{promptcolor}{Constraints: }The output should be `Yes' or `No'.\\*
\\*
\textcolor{promptcolor}{Example 2}\\*
\textcolor{promptcolor}{Instruction: }In this task, you're given a short story of five sentences written in natural language. However, the order of the given story is not correct. Your job is to return the correct order for the given five sentences to create a coherent short story with the new order that has the correct flow. Generate your answer using the number of sentences in the correct order, such as `23415'.\\*
\textcolor{promptcolor}{Input: }Sentence1: He is happy now. Sentence2: Rick grew up in a troubled household. Sentence3: It wasn't long before Rick got shot in a robbery. Sentence4: The incident caused him to turn a new leaf. Sentence5: He never found good support in family, and turned to gangs.\\*
\textcolor{promptcolor}{Constraints: }The output must consist of the numbers representing the sentences.\\*
\\*
\textcolor{promptcolor}{Example 3}\\*
\textcolor{promptcolor}{Instruction: }In this task, you are given Yelp reviews. The task is to classify a review as ``POSITIVE'' if the overall sentiment of the review is positive or as ``NEGATIVE'' if the overall sentiment of the review is negative.\\*
\textcolor{promptcolor}{Input: }This is my go to place to get the best chicken Pad Thai! Also the price is super good, considering most places have high prices and poor quality. Love this place, its definitely a top 5 fav for take out.\\*
\textcolor{promptcolor}{Constraints: }The output should be `POSITIVE' or `NEGATIVE'.\\*
\\*  
\textcolor{promptcolor}{Example 4} \\*
\midrule
\textbf{Seed 3} \\*
\\*
\textcolor{promptcolor}{Example 1}\\*
\textcolor{promptcolor}{Instruction: }You are given a question title. You need to classify it into one of the 10 topics: 0 - `Society \& Culture', 1 - `Science \& Mathematics', 2 - `Health', 3 - `Education \& Reference', 4 - `Computers \& Internet', 5 - `Sports', 6 - `Business \& Finance', 7 - `Entertainment \& Music', 8 - `Family \& Relationships', 9 - 'Politics \& Government'.\\*
\textcolor{promptcolor}{Input :}What's the best way to clean a keyboard?\\*
\textcolor{promptcolor}{Constraints: }The output should be a number between 0 and 9.\\*
\\*
\textcolor{promptcolor}{Example 2}\\*
\textcolor{promptcolor}{Instruction: }You are given a target person and two sentences. Your task is to determine whether the second sentence conveys either a stereotype, anti-stereotype, or is unrelated with respect to the first sentence and the target. A stereotype is a widely held but oversimplified notion of a group of people. An anti-stereotype is an idea that goes against the widely held but oversimplified notion of a group of people. Sentences and targets are unrelated if there is no relevance between them.\\*
\textcolor{promptcolor}{Input: }Target: mother. The mother walked into her kitchen. She pulled a beer out of the fridge and leaned against the doorway, sipping it slowly.\\*
\textcolor{promptcolor}{Constraints: }The output should be one of the three: `stereotype', `anti-stereotype' or `unrelated'.
\\*
\\*
\textcolor{promptcolor}{Example 3}\\*
\textcolor{promptcolor}{Instruction: }In this task, you are given an article. Your task is to summarize the article in a sentence.\\*
\textcolor{promptcolor}{Input: }Ms Bourne, who became Sussex's first PCC in 2012, was declared winner after the vote went to a second count. Three candidates were eliminated in the first count, leaving only Ms Bourne and Labour's Michael Jones. In the first count, Ms Bourne had 114,570 votes and Mr Jones had 61,017. The second count brought their totals to 139,335 and 86,392 respectively. She said:  ``I'm absolutely honoured and very privileged to be elected.'' She said she needed to ``find extra savings because budgets are still reducing'' and ``to invest in frontline policing because I know that is really important to people in Sussex''. Voter turnout was 22.5\% compared with 15.3\% in 2012. The three eliminated in the first count were Green Party candidate James Doyle, UKIP's Patrick Lowe and James Walsh from the Liberal Democrats. Results listed alphabetically by surname are as follows. BBC News App users: tap here to see the results.\\*
\textcolor{promptcolor}{Constraints: }None.\\*
\\*  
\textcolor{promptcolor}{Example 4} \\*
\midrule
\textbf{Seed 4} \\*
\\*
\textcolor{promptcolor}{Example 1}\\*
\textcolor{promptcolor}{Instruction: }In this task, you are given Wikipedia articles on a range of topics as passages and a question from the passage. We ask you to answer the question by classifying the answer as 0 (False) or 1 (True).\\*
\textcolor{promptcolor}{Input: }Passage: Property tax -- Property tax or `house tax' is a local tax on buildings, along with appurtenant land. It is and imposed on the Possessor (not the custodian of property as per 1978, 44th amendment of constitution). It resembles the US-type wealth tax and differs from the excise-type UK rate. The tax power is vested in the states and is delegated to local bodies, specifying the valuation method, rate band, and collection procedures. The tax base is the annual rental value (ARV) or area-based rating. Owner-occupied and other properties not producing rent are assessed on cost and then converted into ARV by applying a percentage of cost, usually four percent. Vacant land is generally exempt. Central government properties are exempt. Instead a `service charge' is permissible under executive order. Properties of foreign missions also enjoy tax exemption without requiring reciprocity. The tax is usually accompanied by service taxes, e.g., water tax, drainage tax, conservancy (sanitation) tax, lighting tax, all using the same tax base. The rate structure is flat on rural (panchayat) properties, but in the urban (municipal) areas it is mildly progressive with about 80\% of assessments falling in the first two brackets. Question: is house tax and property tax are same.\\*
\textcolor{promptcolor}{Constraints: }The output should be 0 or 1.\\*
\\*
\textcolor{promptcolor}{Example 2}\\*
\textcolor{promptcolor}{Instruction: }Rewrite each original sentence in order to make it easier to understand by non-native speakers of English. You can do so by replacing complex words with simpler synonyms (i.e. paraphrasing), deleting unimportant information (i.e. compression), and/or splitting a long complex sentence into several simpler ones. The final simplified sentences need to be grammatical, fluent, and retain the main ideas of their original counterparts without altering their meanings.\\*
\textcolor{promptcolor}{Input: }From its inception, it was designated a duty-free port and vied with the neighboring Sultanate of Pattani for trade.\\*
\textcolor{promptcolor}{Constraints: }None.\\*
\\*
\textcolor{promptcolor}{Example 3}\\*
\textcolor{promptcolor}{Instruction: }You are provided with an arithmetic question. Your task is to compute the solution using the given arithmetic operations. The only arithmetic operators needed to answer the questions are'+'(addition) and'-'(subtraction). The answer should be correct to one decimal place.\\*
\textcolor{promptcolor}{Input: }Joan found 70 seashells on the beach. She gave Sam some of her seashells, after which she has 27 seashell left. How many seashells did she give to Sam?\\*
\textcolor{promptcolor}{Constraints: }None.\\*
\\*  
\textcolor{promptcolor}{Example 4} \\*
\midrule
\textbf{Seed 5} \\*
\\*
\textcolor{promptcolor}{Example 1}\\*
\textcolor{promptcolor}{Instruction: }You are given a science question (easy-level) and four answer options (associated with ``A'', ``B'', ``C'', ``D''). Your task is to find the correct answer based on scientific facts, knowledge, and reasoning. Do not generate anything else apart from one of the following characters: `A', `B, `C', `D'. There is only one correct answer for each question.\\*
\textcolor{promptcolor}{Input: } Which part of a bicycle BEST moves in a circle? (A) Seat (B) Frame (C) Foot pedal (D) Kickstand\\*
\textcolor{promptcolor}{Constraints: }The output should be one of the following characters: `A', `B, `C', `D'.\\*
\\*
\textcolor{promptcolor}{Example 2}\\*
\textcolor{promptcolor}{Instruction: }You are given a negative review and your task is to convert it to a positive review by one or more making minimal changes. Avoid changing the context of the review.\\*
\textcolor{promptcolor}{Input: }we stood there in shock, because we never expected this.\\*
\textcolor{promptcolor}{Constraints: }None.\\*
\\*
\textcolor{promptcolor}{Example 3}\\*
\textcolor{promptcolor}{Instruction: }In this task, you are given two sentences taken from a conversation, and your job is to classify whether these given sentences are sequential or not. We will mark the given sentence pair as `True' if it's sequential, otherwise `False'. The two sentences are spoken by two different people.\\*
\textcolor{promptcolor}{Input: }Noah: When and where are we meeting? :), Madison: I thought you were busy...?\\*
\textcolor{promptcolor}{Constraints: }The output should be `True' or `False'.\\*
\\*  
\textcolor{promptcolor}{Example 4} \\*
\bottomrule
    \caption{The in-context demonstrations used in our experiments.}
    \label{table:incotext-prompts-tab}
\end{longtable}
\endgroup

\twocolumn

\end{document}